\documentclass[lettersize, journal]{IEEEtran}
\usepackage{amsmath, amsfonts}
\usepackage{algorithmic}
\usepackage{algorithm}
\usepackage{array}
\usepackage[caption=false,font=normalsize,labelfont=sf,textfont=sf]{subfig}
\usepackage{textcomp}
\usepackage{stfloats}
\usepackage{url}
\usepackage{verbatim}
\usepackage{graphicx}
\usepackage{cite}
\usepackage{float}
\usepackage{placeins}

\usepackage{textcomp}
\usepackage{stfloats}
\usepackage{url}
\usepackage{verbatim}
\usepackage{graphicx}
\usepackage{booktabs}
\usepackage{filecontents}

\usepackage{lipsum}
\usepackage[table,xcdraw]{xcolor}
\graphicspath{ {./images/} }
\usepackage{balance}

\usepackage{cite}
\usepackage{amsmath,amssymb,amsfonts}
\usepackage{algorithmic}
\usepackage{graphicx}
\usepackage{textcomp}
\usepackage{xcolor}
\usepackage{authblk}

\usepackage{amsmath, amsfonts, amsthm, dsfont, amssymb}

\def\BibTeX{{\rm B\kern-.05em{\sc i\kern-.025em b}\kern-.08em
    T\kern-.1667em\lower.7ex\hbox{E}\kern-.125emX}}
\usepackage{balance}

\begin{document}

\title{NU-Class Net: A Novel Approach for Video Quality Enhancement}

\author[1]{Parham Zilouchian Moghaddam\thanks{*E-mail: p.zilouchian@gmail.com, modarressi@ut.ac.ir, and msadeghi@hbku.edu.qa}}
\author[1]{Mehdi Modarressi}
\author[2]{Mohammad Amin Sadeghi}
\affil[1]{School of Electrical and Computer Engineering, College of Engineering, University of Tehran}
\affil[2]{QCRI / HBKU}



\markboth{IEEE Transactions on Circuits and Systems for Video Technology}%
{How to Use the IEEEtran \LaTeX \ Templates}


\maketitle

\begin{abstract}

Video content has experienced a surge in popularity, asserting its dominance over internet traffic and Internet of Things (IoT) networks. Video compression has long been regarded as the primary means of efficiently managing the substantial multimedia traffic generated by video-capturing devices. Nevertheless, video compression algorithms entail significant computational demands in order to achieve substantial compression ratios. This complexity presents a formidable challenge when implementing efficient video coding standards in resource-constrained embedded systems, such as IoT edge node cameras. To tackle this challenge, this paper introduces NU-Class Net, an innovative deep-learning model designed to mitigate compression artifacts stemming from lossy compression codecs. This enhancement significantly elevates the perceptible quality of low-bit-rate videos. By employing the NU-Class Net, the video encoder within the video-capturing node can reduce output quality, thereby generating low-bit-rate videos and effectively curtailing both computation and bandwidth requirements at the edge. On the decoder side, which is typically less encumbered by resource limitations, NU-Class Net is applied after the video decoder to compensate for artifacts and approximate the quality of the original video. Experimental results affirm the efficacy of the proposed model in enhancing the perceptible quality of videos, especially those streamed at low bit rates.

\end{abstract}


\section{Introduction}
\IEEEPARstart{T}{he} boost in video content usage, driven by a demand for superior quality, has significantly amplified storage and network bandwidth needs in recent years. A study by Cisco reveals that video content comprises over 80\% of global Internet traffic\cite{cisco2020cisco}. This figure unexpectedly soared during the COVID-19 pandemic, as work, education, and leisure increasingly pivoted to live video platforms. Furthermore, with numerous Internet-of-Things (IoT) services and applications incorporating video transmission\cite{survey22}\cite{chen2020internet}, managing video traffic has become a pivotal concern in contemporary IoT systems\cite{iot-ref}.

The prodigious volume of video content not only demands substantial storage space and network bandwidth but also contributes to an estimated 3.7\% of the CO2 emissions generated by the web\cite{bbcfuture}. This issue becomes particularly salient in the current context, where the energy consumption of computing systems and its subsequent impact on environmental and climate change has elicited widespread concern. Consequently, even a marginal reduction in video size can significantly curtail bandwidth usage, storage space, and associated energy consumption. This underscores the pivotal role of video compression in facilitating the energy-efficient storage and transfer of video content.

Video compression, or coding, endeavors to diminish video size by excising statistical, temporal, and spatial redundancies within frames and, to a degree, omitting non-critical details such that the resultant quality degradation remains imperceptible to human vision.

Inherent in video streaming is a delicate balance between perceptible video quality and bit rate: an elevation in video quality invariably necessitates a proportional increase in bit rate and, consequently, bandwidth demand. Cutting-edge compression methodologies employ a myriad of sophisticated algorithms to navigate the trade-off between quality and bandwidth utilization judiciously. Nonetheless, achieving a symbiosis of low bit rate and high video quality invariably precipitates a surge in computational complexity on both the encoder and decoder fronts. For instance, Versatile Video Coding (VVC), one of the most recent MPEG standards, leverages a multitude of coding tools and AI-based algorithms to realize a 25-50\% bit-rate reduction while maintaining quality comparable to preceding coding standards. This, however, escalates encoder and decoder complexities by factors of up to 27\texttimes \space and 2\texttimes, respectively\cite{chen2019algorithm}.

Energy consumption in encoding and decoding processes frequently escalates in tandem with computational complexity. Such elevated complexity and energy expenditure hinder the implementation of high-quality video coding standards across numerous embedded devices and IoT edge nodes, such as cameras mounted on battery-operated drones, which typically operate under stringent energy budgets and possess limited computational capabilities. Indeed, video coding in such systems must adeptly navigate the intricate balance among three competing demands: quality, bit rate, and complexity.

In this paper, we introduce UN-class, a comprehensive deep learning-based technique devised to navigate the challenges of video encoding amidst constrained resources and limited network bandwidth. The nomenclature `NU-Class' was inspired by its semblance to a space shuttle featured in the Star Wars series. Employing this technique, the encoder intentionally moderates video quality constraints, generating a low-bit-rate video albeit with a consequential quality degradation. Conversely, on the decoder's side, a deep learning model is utilized. This model, which is meticulously trained to counterbalance the quality loss by mitigating coding artifacts, endeavors to reconstruct the video, aspiring to attain a quality proximate to the original.

A salient feature of NU-Class Net distinguishes it from numerous contemporary works employing deep learning to devise new codecs or enhance existing ones: it does not modify or supplant a codec. Rather, it amplifies the end-to-end coding process by integrating a deep learning module subsequent to the decoder, with the aim of elevating the quality of the decoded video. Consequently, UN-Class maintains orthogonality and can be synergistically utilized with any modern video codec to mitigate the video stream bit rate further.

NU-Class Net facilitates a less resource-intensive video coding at the encoder side by involving a simplified encoding process, concurrently reducing the network bandwidth necessary for video stream transmission. On the decoder side, a deep learning model is strategically positioned subsequent to the decoder module to reconstruct video quality. The advantages of UN-Class in diminishing bandwidth and storage requisites are conspicuous. It proves exceptionally advantageous for embedded systems and Internet of Things (IoT) nodes tasked with capturing and transmitting videos under stringent energy and processing constraints. With UN-Class, these devices enjoy the luxury of less complex video encoding, obviating the need to operate intricate encoder modules that enhance quality. Correspondingly, the energy consumption for communication diminishes in proportion to the bit rate reduction. In such systems, the complexity is translocated to the decoder side, typically operating on a computer characterized by a more lenient power budget and a robust processing unit.

NU-Class Net is architecturally founded upon the U-Net, a preeminent AutoEncoder deep learning model. AutoEncoders have catalyzed breakthroughs across various computer vision tasks, especially in the realm of image reconstruction. The advent of U-Net has markedly outperformed competitors and transformed the application of encoder-decoder architectures across diverse computer vision tasks. Given that videos are composed of frames, which can be regarded as images, an opportune avenue emerges to tailor U-Net for image enhancement. NU-Class Net ingests compressed video frames as input and predicts the residual difference between the original (attainable through high-quality encoding) and compressed (low bit-rate) frames. This residual is superimposed upon the input frames to alleviate discernible encoding artifacts, thereby maintaining an output video quality that approximates the original.

While there exists a modicum of prior work focused on enhancing the quality of JPEG images via neural networks, NU-Class Net is meticulously designed with a specific emphasis on video, harnessing the inter-frame correlation inherent in video streams to augment performance\cite{ledig2017photo}\cite{maleki2018blockcnn}. Moreover, NU-Class Net exclusively computes the residual difference between raw and compressed frames, eschewing the generation of an entire frame. This approach strategically circumvents the arduous task of predicting an entire image with its intricate details, thereby enhancing the network’s efficiency and expediting its training. Our results substantiate that NU-Class Net significantly elevates the quality of low-bit-rate video frames, effectively empowering video capturing nodes to utilize low-complexity, low-bit-rate video streams.

The remainder of this paper is structured as follows: Section 2 provides a review of pertinent related work. Section 3 introduces the architecture of NU-Class Net, while Section 4 delineates the implementation details. Evaluation results are presented in Section 5, and Section 6 offers concluding remarks.

\section{Related Work}
\noindent Employing deep learning for video stream compression can be approached via various methodologies, which can be broadly classified into three primary categories as follows.

\subsection{Integrating deep learning modules into codecs}
\noindent One approach to augmenting various modules of existing video codecs involves the utilization of deep learning-based models. For instance, Golinski et al. \cite{golinski2020feedback} introduce a novel deep-learning architecture that adeptly learns video compression in low latency mode, also discussing temporal consistency issues encountered during their experiments and proposing solutions thereto. Alternatively, Pourreza et al. \cite{pourreza2021extending} propose a neural video codec capable of managing B-frame coding predicting a frame from both future and past reference frames as opposed to the more prevalent P-frame coding \cite{majumdar2004distributed} utilized by the majority of neural encoders. This method interpolates future and past reference frames to derive a single frame, which is then employed with an existing P-frame codec, seamlessly integrating with current neural codecs that predominantly rely on the P-frame predictor. Rozendaal et al. \cite{van2021instance} enhance existing neural codecs using a technique termed instance-adaptive learning, wherein they seek to modify a pre-trained compression model to transmit optimal parameters to the receiver alongside the latent code.

A predominant drawback of the aforementioned methods resides in their lack of portability; specifically, both the sender and receiver must be equipped with codecs that are deep learning-enhanced. In contrast, our approach permits the utilization of any standard codec on both ends, simply appending a neural network module subsequent to the decoder to enhance quality.

\subsection{Image enhancement using deep learning}
In this approach, a neural network is trained for end-to-end data compression, operating in conjunction with the codec. For instance, Maleki \emph{et al.} introduced Block CNN, a novel technique designed to eliminate JPEG coding artifacts \cite{maleki2018blockcnn}. The image is partitioned into 8\texttimes8 pixel blocks, with the intensities of each block predicted based on preceding blocks, followed by the computation of the input image's residual. This technique repurposes JPEG’s legacy processes for compression and decompression, achieving superior results compared to baselines at high compression ratios. Contrasting with Block CNN, NU-Class Net concentrates on video frames and aspires to capture lower-level features, distinguishing our work from theirs.

Ronneberger \emph{et al.} pioneered the development of the U-shaped network, U-Net, designed explicitly for semantic segmentation\cite{ronneberger2015u}. The algorithm astutely abstracts compact features using a bottom-up approach, subsequently mapping them back to the original scale through a top-down branch. A notable enhancement over standard autoencoders, U-Net employs skip connections to amalgamate low-level and high-level features. The model's stellar performance has not only set a benchmark but also spawned the development of several derivatives, adept at managing more intricate image segmentation and generation tasks.

Vaccaro \emph{et al.} introduced SR-UNet, a tailored U-Net model, aimed at real-time enhancement of decoded video quality directly on the user's device\cite{vaccaro2021fast}. This neural network adeptly undertakes super-resolution—reconstructing high-resolution frames from a low-resolution stream—and mitigates artifacts emerging post-video compression. Employed as a final post-processing stage, SR-UNet elevates the visual fidelity of a frame prior to its display, facilitating the utilization of existing video coding and transmission pipelines without necessitating modifications. It's pertinent to note, however, that while effective, these methods inherently alter image resolution and, consequently, video size—an outcome not always desirable. Contrarily, our approach abstains from modifying resolution, placing a focused effort on artifact reduction.

In \cite{noura_2021}, they address the challenge of transmitting high-quality multimedia content in resource-limited MIoT environments. The authors propose a deep learning-based approach, utilizing a Residual Dense Network (RDN) for image super-resolution. This method significantly improves the visual quality of heavily compressed images transmitted from MIoT devices, thus optimizing both communication overhead and power consumption.



\subsection{Exploring Video Synthesis through Generative Models}
Recent years have witnessed the evolution of models employing residual learning, all aiming to attain superior image quality\cite{he2016deep}. However, a direct consequence of increasing the depth of these models\cite{zhang2018image}, \cite{lim2017enhanced} has been a notable surge in computational complexity and memory usage.



Numerous studies exploring the use of Generative Models in video compression have predominantly focused on video encoding. Remarkably, some of the most promising results have emerged from utilizing Generative Adversarial Networks (GANs) to produce photo-realistic videos from sequences of semantic maps. Take, for instance, the Fast-Vid2Vid experiment by Long Zhuo et al.\cite{zhuo2022fast}, which synthesizes photo-realistic video from semantic maps leveraging GANs. Contrasting with our approach, these methods predominantly zero in on generating entirely new videos from scratch, rather than enhancing the quality of existing footage.

\section{Proposed Method}
\subsection{System Design}
\noindent Contrary to conventional approaches that primarily focus on optimizing and modifying the codec itself, we propose a nuanced solution that emphasizes enhancing the frames post-codec processing. This method diverges from traditional strategies that attempt to integrate artificial intelligence (AI) modules directly into the codec, opting instead to concentrate on improving frame quality subsequent to the codec's operation.

\begin{figure}[!h]
\centering
\includegraphics[width=3.7in]{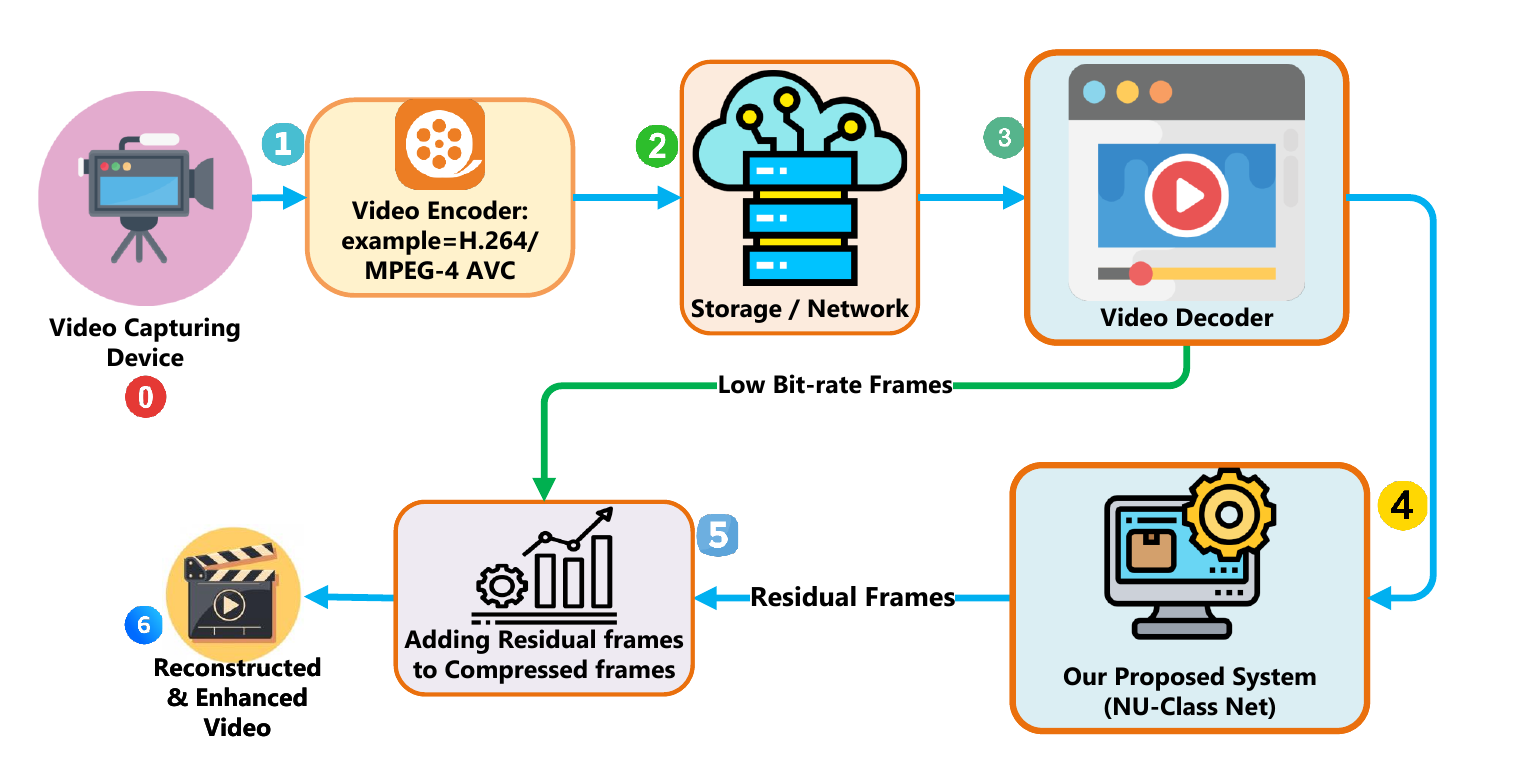}
\caption{Schematic Representation of the Comprehensive Application Process of the NU-Class Model}
\label{Goal_System}
\end{figure}


Fig.\ref{Goal_System} delineates the overarching design of our system, employing the NU-Class Net model to enhance the input video quality. This technique affords a notable opportunity to compress video beyond the capacities of conventional codecs, such as HEVC\cite{6316136}. A salient advantage of our proposed method lies in its versatility; it can be appended as an extension to any video codec, irrespective of its underlying technology.

As the network depth augments and additional pooling layers are utilized to reduce the input image size, there is a consequential loss of lower-level features—critical, especially for our task. To counteract these challenges, we introduce a bespoke model designed to enhance imported video quality. We have christened our model 'NU-Class Net,' drawing inspiration from a space shuttle featured in the Star Wars series. The nomenclature arises from a visual resemblance: our proposed network architecture mirrors the general shape of this specific space shuttle.

\begin{figure}[!h]
\centering
\includegraphics[width=3.5in]{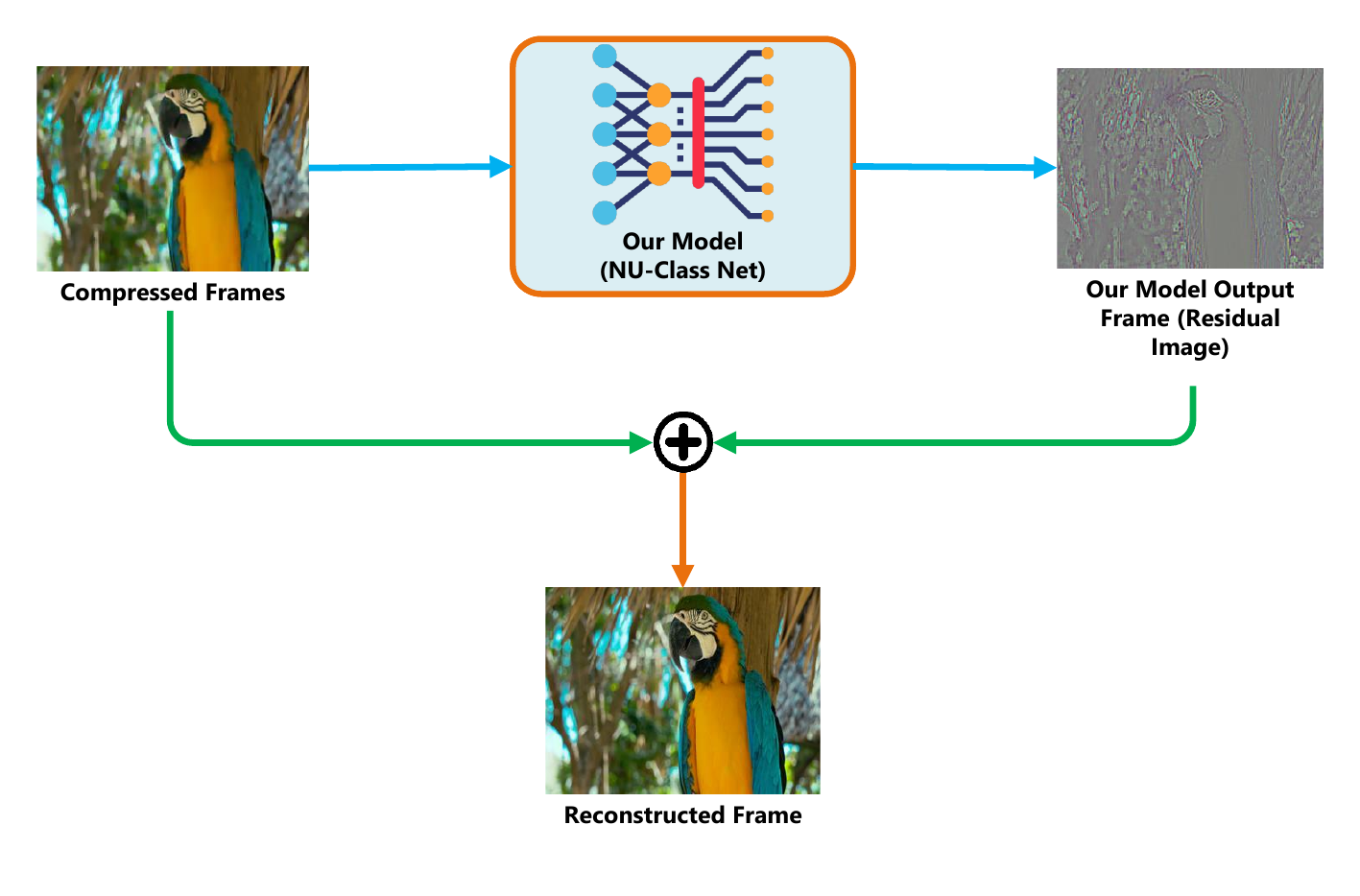}
\caption{How Base NU-Class Net performs.}
\label{Base_NU_Class_Net}
\end{figure}


The NU-Class Net processes frames at reduced bit rates, received from the video encoder, and produces what we designate as the 'residual frame' for a given input image. The residual frame embodies the discrepancy between the video frame—at its reduced bit rate—and its unaltered counterpart, maintaining equivalent resolution and aspect ratio. Upon generation of the residual frame, it is added to the input image, which exhibits a reduced bit rate. Consequently, the final image output closely resembles the original, as depicted in Fig.\ref{Base_NU_Class_Net}. A cardinal advantage of our method is the substantial reduction in video file size, achieved without altering the foundational video codec or its associated modules.

\subsection{Architectural Design of the NU-Class Net Model}
\begin{figure*}[!t]
\centering
\includegraphics[width=6.5in]{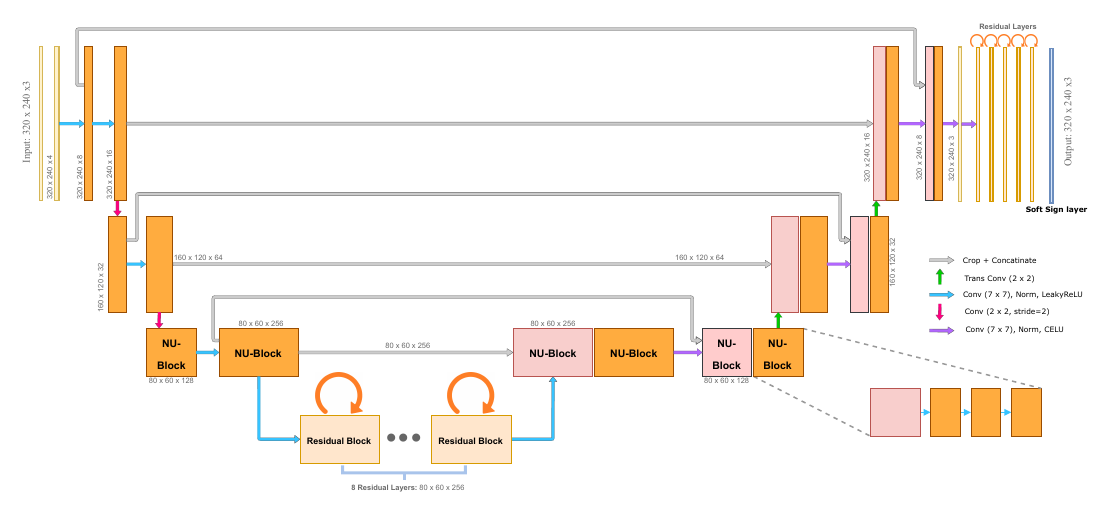}
\caption{Detailed Architecture of the NU-Class Net, Illustrating Layers, Nodes, and Connectivity Patterns.}
\label{NU_Class_Net}
\end{figure*}

\noindent Fig.\ref{NU_Class_Net} outlines the architecture of NU-Class Net, embodying an encoder-decoder design paradigm. Contrary to focusing on the extraction of pivotal features for tasks like classification, our foremost objective is to derive high-level features of a video to facilitate quality enhancement. In devising the architecture of NU-Class Net, we drew inspiration from the U-Net architecture to aptly cater to this prerequisite\cite{ronneberger2015u}.

\begin{figure}[!h]
\centering
\includegraphics[width=3.5in]{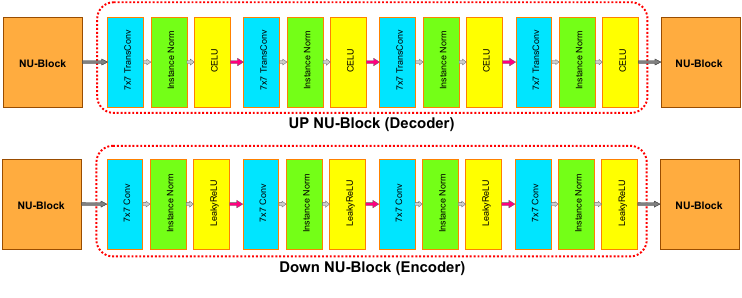}
\caption{Internal Composition of the NU-Block, Featuring Sub-components within Decoder and Encoder Structures.}
\label{NU_Block}
\end{figure}

\begin{figure}[!h]
\centering
\includegraphics[width=3.5in]{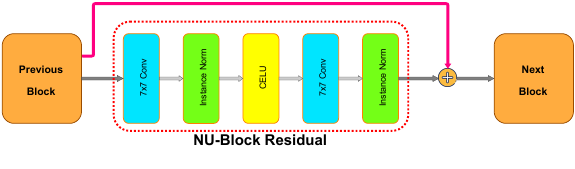}
\caption{Schematic Illustration of the Inner Structure and Operational Flow within the NU-Block Residual.}
\label{NU_Class_Residual}
\end{figure}

Our proposed architecture pivots around five crucial components, each integral to the enhanced functionality and performance of the NU-Class Net:

\begin{itemize}
\item{{\bf{Encoder blocks:}} The encoder section of our model is constructed of six distinct blocks, each encompassing four convolution layers. Notably, two blocks integrate pooling layers, for which we eschew traditional pooling methods in favor of convolutions with a stride of two, enhancing computational efficiency. Furthermore, we have expanded the receptive field of these blocks beyond the conventional 3\texttimes3 to 7\texttimes7 convolution windows, optimizing feature extraction (refer to Fig.\ref{NU_Block} for the structure of each encoder block).
}
\item{{\bf{Bottleneck residual blocks (or ResBox or Residual connections):}} The bottleneck segment of our network thoughtfully integrates eight residual blocks, which are instrumental in enabling the model to learn identity functions and thereby facilitate the training of a deeper neural network without incurring a consequential decrement in performance. This is actualized by adeptly mitigating issues related to both vanishing and exploding gradients\cite{luo2016understanding}. The formulation of a residual block is presented as (\ref{eq1}). The structure of each implemented residual block is illustrated in Fig.\ref{NU_Class_Residual}.
\begin{equation} \label{eq1}
{G(x)}={F(x)}+x
\end{equation}
}
\item{{\bf{Decoder blocks:}} The decoder blocks symmetrically counterpoint the processes of the encoder blocks, albeit in a reversed sequence. A consistent filter size is meticulously maintained throughout the architecture. Opting for a different approach than utilized in the encoder blocks, transpose convolutions are employed in place of standard convolutions and pooling. A detailed configuration of each decoder block is illustrated in Fig.\ref{NU_Block}).
}
\item{{\bf{Skip-connections:}} Skip-connections, bridging the Encoder and Decoder blocks, enable the neural network to harness both high-resolution and low-level feature information, capturing meticulous details at every pixel position—paramount for our application. This detailed information is directly conveyed to a latent layer within our network, safeguarding this invaluable data. Consequently, subsequent layers in our network are endowed with both lower resolution, high-level spatial, and contextual information, as well as low-level, texture-like information, rich in detail.
}
\item{{\bf{Final residual blocks:}} Ultimately, five additional residual blocks are strategically positioned at the terminal point of our architecture, acting as guardians and conduits of indispensable information. These blocks not only preserve and transmit the most valuable data necessary for our application but also significantly mitigate the blurriness of the reconstructed frames, ensuring a sharp and clear output.
}
\end{itemize}

Leveraging skip-connections, originally proposed in U-Net, facilitates the extraction of higher-level and quintessential features from the data. The Encoder segment of our model is composed of six Contracting Blocks, each housing four convolution blocks. Notably, each Contracting Block is linked to the Expanding Blocks in the Decoder segment via skip-connections. The Decoder, effectively a mirrored replica of the Encoder, incorporates two inputs across its six layers: 1) The output from the preceding layer and 2) A skip connection from the corresponding block in the Encoder. Mirroring the Encoder, the Expanding Blocks within the Decoder are comprised of four convolution layers identical to those in the Contracting Block, ensuring a symmetrical and cohesive architecture.

In regards to the normalization layers, instance normalization\cite{ulyanov2016instance} was employed post each convolution layer within our network. This technique normalizes across each channel in individual training examples as opposed to normalizing across input features within a single example. Our preference, for instance, normalization, over alternatives such as batch or group normalization, is principally driven by our objective to predict the information dissipated across each channel or the imposed noise. Consequently, instance normalization emerges as a particularly advantageous strategy for our application, aligning closely with our specific concerns and operational requirements.

It merits particular note that two Feature Map Blocks have been integrated and positioned strategically at both the inception and conclusion of our entire network. These blocks shoulder the responsibility of emphasizing the convolution channels to desired channels. For example, they facilitate the modification of channel numbers, such as transmuting a four-channel input to a three-channel RGB output or vice versa, ensuring adaptability and relevance in both the input and output phases of our model.

We strategically augmented the kernel size of the convolution windows from 3\texttimes3 to 7\texttimes7, a decision steered by the necessity to capture an expansive array of information and precise shapes to bolster their quality. Small filter sizes, such as 3\texttimes3, emerged as suboptimal for our particular use case, underlining the imperative to meticulously govern the receptive field, since elements outside of it will not influence the value of that unit. Our motivation pivots on ensuring comprehensive coverage of the entire pertinent area within the frame region. Achieving accurate prediction of every individual pixel in a given image (especially pivotal for dense prediction tasks like image segmentation, optical flow estimation, and our project, which endeavors to predict every pixel in the input image) demands that each output pixel maintains a relatively substantial receptive field to preclude the omission of critical information\cite{luo2016understanding}. Given that our project is geared towards the enhancement of pixel and video frame quality, it navigates through lower-level features in the extracted frames. Additionally, the incorporation of a relatively extensive receptive field in our crafted model empowers us to capture the lower-level features pervading our dataset more precisely.

\subsection{Sequential NU-Class Net}
\begin{figure}[!h]
\centering
\includegraphics[width=3.5in]{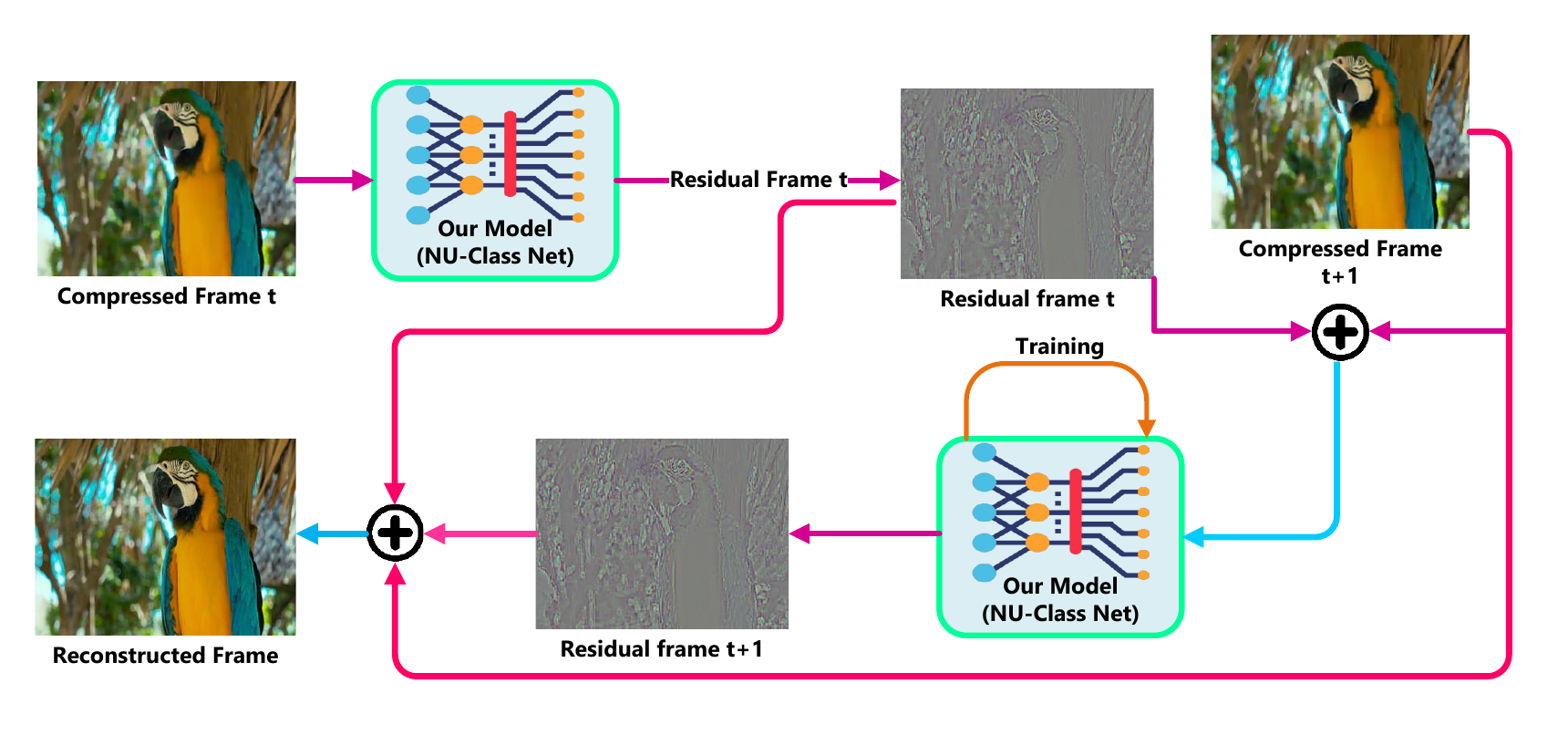}
\caption{Architecture and Training Workflow of the Sequential NU-Class Net Model.}
\label{Sequential_NU_Class_Net}
\end{figure}

\noindent To address the inherent challenges of optical flow and frame consistency (imperative elements in video processing), we have integrated a novel feature into our model, denominated the ``NU-Class Net Sequential." This feature enables us to impart more precise temporal consistency while simultaneously minimizing flickering artifacts between frames. The architecture of this methodology is visually articulated in Fig. \ref{Sequential_NU_Class_Net}, elucidating our strategy of utilizing the residual results from preceding frames in subsequent ones. While this innovative approach to employing previous residuals can subtly enhance results in certain videos, particularly those characterized by action scenes and rapid frame transitions, it is not without potential pitfalls. Despite its capacity to marginally elevate the quality of the resultant video, it is the method’s training speed that notably distinguishes it from its counterparts. Utilizing this methodology requires up to 30\% fewer epochs to attain the optimal state and adequately train the network. The cornerstone of this achievement lies in leveraging knowledge gleaned from preceding frames to predict succeeding ones.

During the training phase, the output of the model for frame t is added to the compressed frame t+1. This combined frame is then input into the model to predict its residual frame, serving as the basis upon which the model is trained. Notably, while the process remains consistent during the inference phase, the model ceases to be trained.

\subsection{Diffusion NU-Class Net}
\begin{figure*}[!t]
\centering
\includegraphics[width=6.5in]{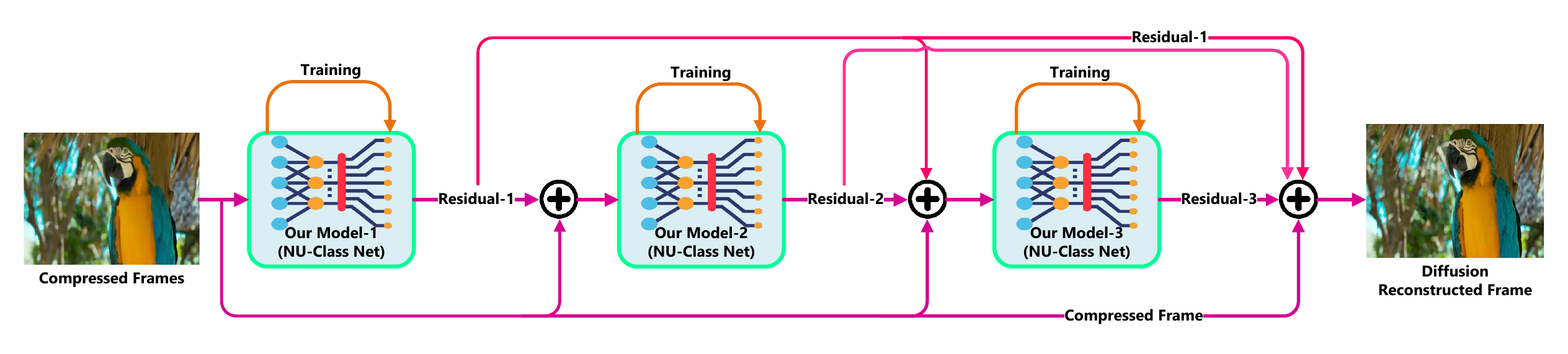}
\caption{Visualization of the Training Process and Practical Implementation of the Diffusion NU-Class Net, Rooted in Diffusion Model Principles.}
\label{diffusion_NU_Class_Net}
\end{figure*}

\noindent Leveraging the core principles of diffusion models\cite{song2020score}, we introduce the Diffusion NU-Class Net. This approach operationalizes the foundational concept of diffusion models, which systematically eliminate added noise from the image through iterative processes. To predict the frame residual, we employ three NU-Class Net models, executing the prediction in a three-stage process. The selection of 'three' emerges as a strategic compromise between optimizing model runtime and enhancing result quality. A detailed depiction of the network architecture and its training methodology is presented in Fig. \ref{diffusion_NU_Class_Net}.

During the training phase, a sequential training approach is implemented, initiating with the first model. Upon its completion, the trained first model serves as a foundation for training the subsequent model. Subsequently, both the first and second trained models collaboratively facilitate the training of the third model. For inference, all three trained models are synergistically utilized in a series, progressively enhancing the input frame quality at each stage.

\subsection{Exploring Generalizability in NU-Class Net}
\noindent The hallmark of our approach, distinguishing it from other methods, lies in its notable generalization capability: NU-Class Net proficiently performs not only on the videos it has been trained on but also on unseen ones. This unique feature stems from the model’s focus on learning detailed, fine-grained lower-level features of video frames instead of concentrating on high-level contextual information. Consequently, the model maintains satisfactory performance even with unfamiliar videos, owing to its utilization of lower-level features and its foundational design for the overarching task of enhancing unseen video quality. It is crucial to note that the model’s generalization performance is substantially influenced by the training dataset and the diversity of contents encompassed within it.

\section{Implementation}
\noindent In the implementation of our approach, images of dimensions 240\texttimes320 were utilized as video frames. The network was trained employing NVIDIA GPUs, specifically the Tesla P100 (16GB)\cite{nvidiap100}, V100 (16GB)\cite{nvidiav100}, and A100 (40GB)\cite{nvidiaa100}, within the Google Colab Pro and Pro+ environments. The training was conducted over a span of 200 epochs to ensure robust learning and convergence. Our model, comprising 79,975,939 parameters, is structured with 12 NU-Blocks and 13 NU-Block Residuals, meticulously designed to balance complexity and computational efficiency.

To expedite and enhance the training process, we leveraged schedulers for learning rate adjustment, enabling a dynamic reduction in the learning rate contingent upon specified validation measurements, and applied post-optimizer update. The “ReduceLROnPlateau” scheduler was utilized, which judiciously diminishes the learning rate upon detecting a plateau in the chosen performance metric, herein, the validation loss. This strategy is intended to judiciously modify the learning rate in alignment with the observed stagnation in performance metrics, facilitating a more nuanced adaptation throughout the learning phase.

In the model’s learning process, Adam \cite{kingma2014adam}, was employed as the optimizer, renowned for its efficacy in various deep learning applications. Furthermore, the loss function utilized for our model was Pixel-Distance Loss \cite{isola2017image}, which leverages the Mean Absolute Error (MAE, L1Loss) metric. The formulations for MAE, Mean Squared Error (MSE), and Pixel Loss are provided as Equations (\ref{eq2}), (\ref{eq3}), and (\ref{eq4}), respectively.

\vspace{-3mm}
\begin{equation} \label{eq2}
{\ell(x, y)}={{\{l_{1}, …, l_{N}\}}^{T}}, \; {l_{n}}={\mid x_{n}-y_{n}\mid} \; (MAE\:Loss)
\end{equation}
\vspace{-3mm}
\begin{equation} \label{eq3}
{\ell(x, y)}={{\{l_{1}, …, l_{N}\}}^{T}}, \; {l_{n}}={(x_{n}-y_{n})^{2}} \; (MSE\:Loss)
\end{equation}
\vspace{-3mm}
\begin{equation} \label{eq4}
{\ell(x, y)}={{\{l_{1}, …, l_{N}\}}^{T}} = \lambda * \Sigma{\mid x_{n}-y_{n}\mid} \; (Pixel\:Loss)
\end{equation}

We opted for this particular loss function over Mean Squared Error Loss (MSELoss) due to the pixel-wise differences under examination; each of these differences is less than one. The Mean Absolute Error (MAE) proves superior in capturing these discrepancies compared to the Mean Squared Error (MSE) as it computes the average absolute difference, providing a measure that is more robust to outliers. Consequently, MAE facilitates a more adept handling of minor variations, thereby optimizing the utility of our model by adeptly managing nuanced differences.

A noteworthy aspect of our approach to loss calculation involves utilizing "Residual Loss" as opposed to conventional loss metrics. This method seeks to compute the disparity between the output of our model and the Pixel Distance Loss incurred between raw and compressed frames. This innovative strategy enables our model to predict the residual of the compressed frames rather than the frames themselves. Subsequently, this predicted residual is added to the compressed frame, thereby enhancing its quality. This technique proficiently addresses challenges encountered when processing video frames, such as rapid scene alterations and diverse video scene compositions, by circumventing issues potentially arising from them.

\vspace{-0mm}
\begin{figure*}[!t]
\centering
\includegraphics[width=6.5in]{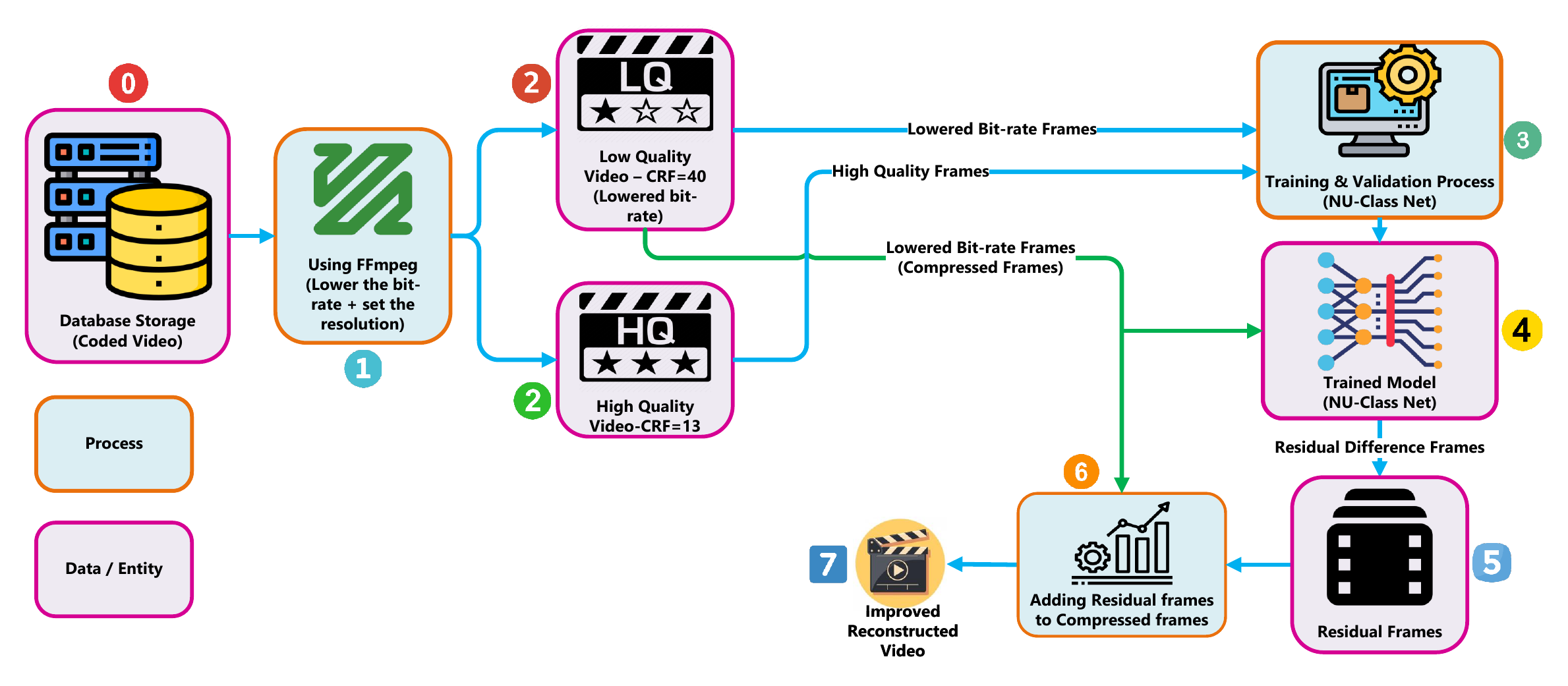}
\caption{Schematic Overview of the Seven-Stage Training Process Applied to All Three Variants of the NU-Class Net.}
\label{fig7}
\end{figure*}
\vspace{-0mm}

A pivotal aspect warranting discussion pertains to the dataset utilized in our study. Given the absence of a pre-existing dataset specifically tailored to this distinctive task and devoid of analogous applications, we curated our own. This was accomplished by utilizing a video file approximately 2 hours and 15 minutes in length, with a frame rate of 30 FPS. Subsequently, frame extraction was performed using the ubiquitously-employed FFmpeg tool \cite{ffmpeg}\cite{tomar2006converting}, which affords the capability of extracting video frames at any designated frame rate. For the construction of our dataset, six frames per second were selected. These frames were extracted under two distinct conditions: low quality and high quality. Within FFmpeg, the Constant Rate Factor (CRF) value determines the quality of the output video, with a lower CRF yielding higher quality albeit at an elevated bit rate. A CRF value ranging from 17 to 23 is conventionally deemed a judicious balance between quality and rate. In our work, we elected to use CRF values of 13 and 40 for raw and low-bit-rate videos, respectively.

Approximately 15 minutes of our video were allocated for testing purposes, with the remainder being dedicated to training.

\section{Experiments \& Results}
\noindent In this section, we critically evaluate the performance of the proposed network, employing three distinct metrics: Pixel-Distance Loss, Peak Signal-to-Noise Ratio (PSNR), and Structural Similarity Index Measure (SSIM). Subsequent discussions will compare and interpret the derived results, providing a comprehensive analysis of the network's efficacy and robustness.

\begin{figure}[!h]
\centering
\includegraphics[width=3.5in]{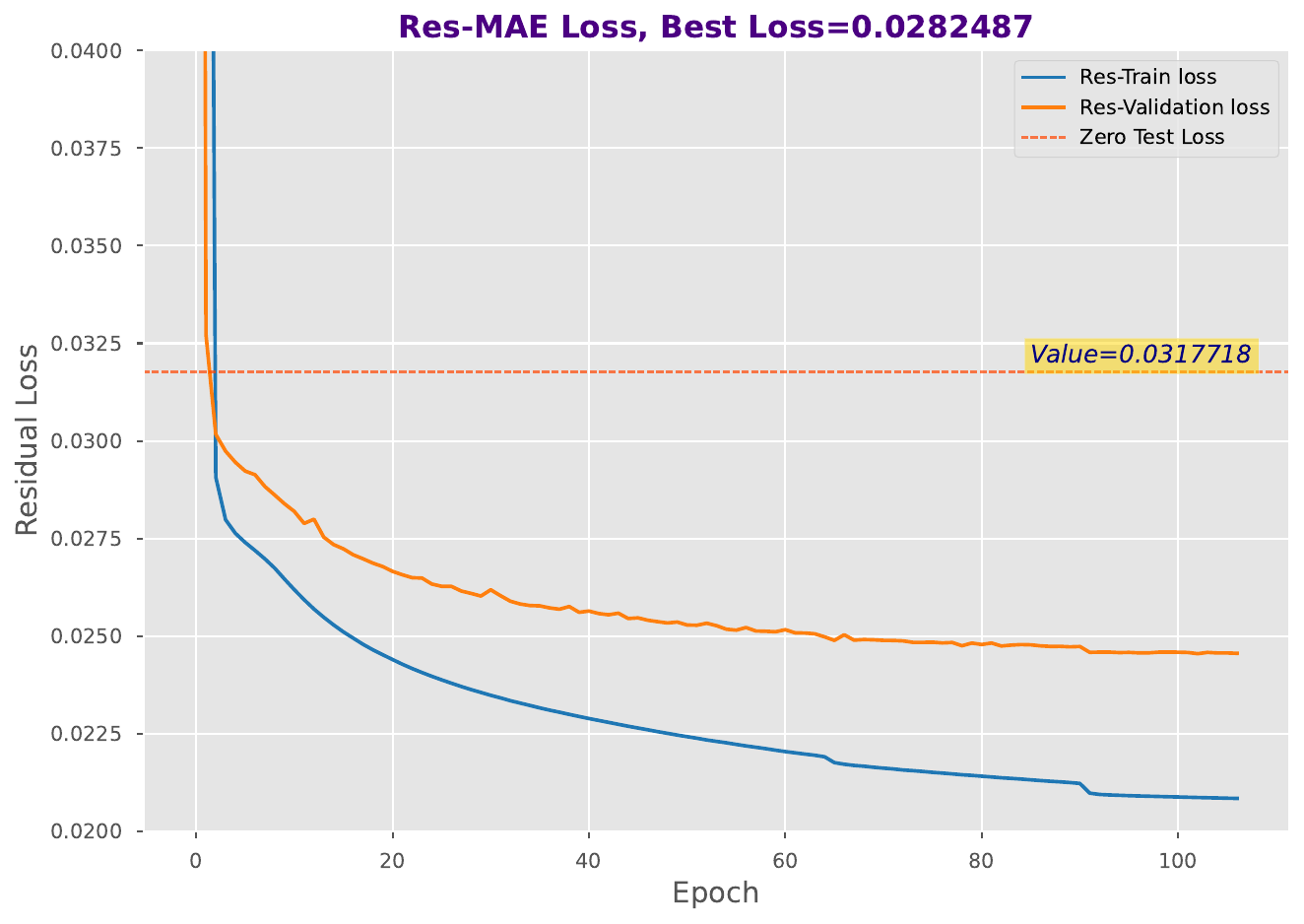}
\caption{Basic NU-Class Net Train and Test calculated loss for each epoch.}
\label{MAE Loss}
\end{figure}

Initially, we delve into the results derived from the Pixel-Distance Loss metric. As illustrated in Fig. \ref{MAE Loss}, a conspicuous enhancement in the quality of the reduced bit-rate frames is achieved using the NU-Class Net.

\begin{figure}[!h]
\centering
\includegraphics[width=3.5in]{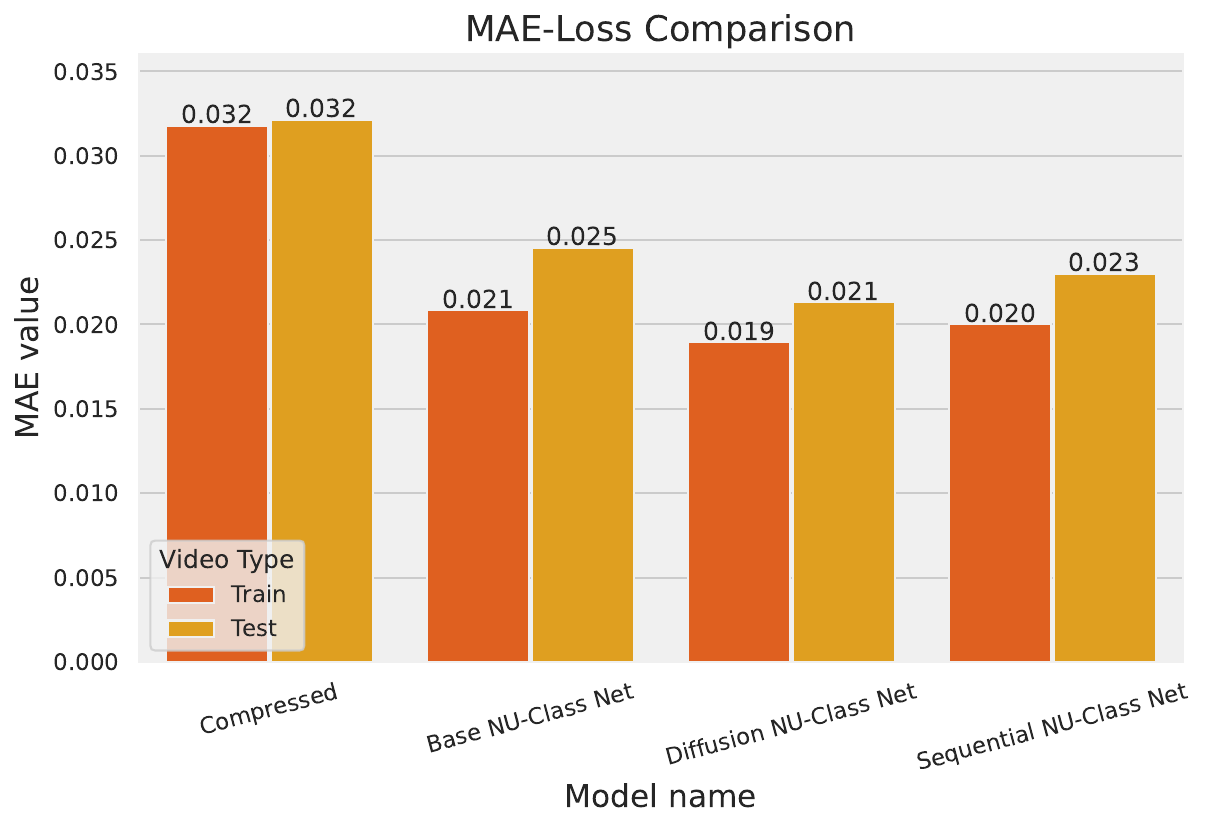}
\caption{The final MAE loss comparison.}
\label{MAE_Loss_Comparison}
\end{figure}
\begin{figure*}[hbt!]
\centering
\includegraphics[width=6.5in]{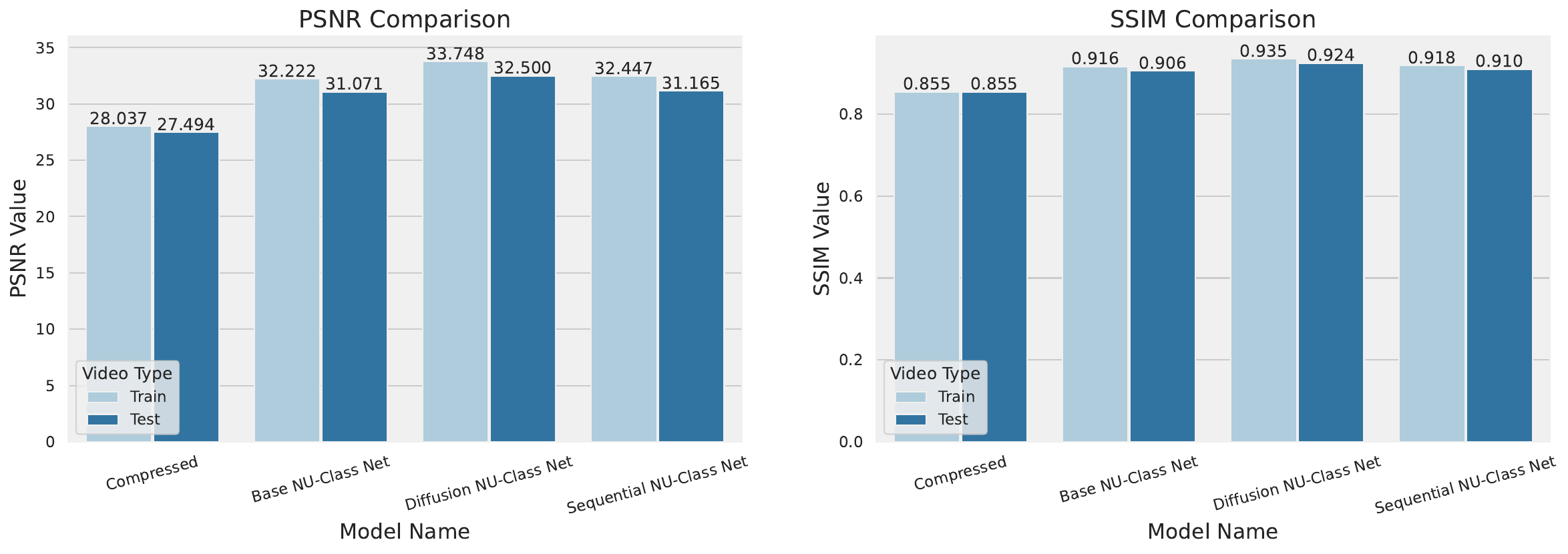}
\caption{Comparative Analysis of PSNR and SSIM Values Across All Proposed Models, Distinguished Between Train and Test Frames.}
\label{PSNR_SSIM}
\end{figure*}

Centering our attention on the Pixel Distance Loss, this metric elucidates the disparities between pixels within every corresponding frame in the dataset. Moreover, as delineated in Fig. \ref{MAE Loss}, there is a discernible reduction in the loss of the test set, approximately 30 percent, upon the conclusion of the training process.

\begin{table*}[!h]
\begin{center}
\caption{Comparative Properties of Raw and Compressed Videos, Including Metrics such as Bitrate, Frame Rate, and Resolution.}
\label{tab1}
\setlength{\arrayrulewidth}{0.15mm}
\setlength{\tabcolsep}{18pt}
\renewcommand{\arraystretch}{1.5}
\begin{tabular}{|l | c c c c c c|}
\hline\hline
\rowcolor[HTML]{EFEFEF} 
                         Video Type & CRF & Bit Rate & Video Resolution & Color Space & Bit Depth & Video File Size \\ \hline
\cellcolor[HTML]{C0C0C0} Raw Video & 13 & 3572 kb/s & 320\texttimes240 & RGB & 8-bit & 7910 MB  \\ \hline
\cellcolor[HTML]{C0C0C0} Compressed Video & 40 & 235 kb/s & 320\texttimes240 & RGB & 8-bit & 445.7 MB\\ \hline
\cellcolor[HTML]{C0C0C0} Typical Video & 18 & 1947 kb/s & 320\texttimes240 & RGB & 8-bit & 4652.6 MB  \\ 
\hline\hline
\end{tabular}
\renewcommand{\arraystretch}{1}
\end{center}
\end{table*}


Table \ref{tab1} delineates the attributes of both the compressed, raw videos utilized in our model, with a chosen resolution of 320\texttimes240. This resolution was selected for two pivotal reasons: initially, the resource requirements for training our model with larger frames exceeded our available resources. Subsequently, the potential storage demands for larger resolutions posed a conceivable challenge. It is imperative to note that, given the proof-of-concept nature of our endeavor, the resolution does not assume a critical role in validating our proposed approach.

Furthermore, as denoted by Table \ref{tab1}, we have achieved a reduction in the video file size by a factor of up to seventeen in the compressed version, while maintaining the original video resolution. This implies that, with identical resolution and aspect ratio, our compressed video occupies only 1/17th of the hard disk space used by the original. This reduced space requirement allows for the storage of additional content within the same capacity without compromising initial quality. Given that our model is designed to enhance the quality of the compressed video, an improvement in video quality is an anticipated outcome.

Moreover, Table \ref{tab1} presents the characteristics of what we define as a 'typical video,' a video that experiences some degree of lossiness without significantly impacting the user experience. When comparing the attributes of a compressed video to this typical benchmark, it is evident that our method achieves an approximate tenfold reduction in size. This significant compression rate is maintained even under conditions where video quality is paramount, demonstrating the effectiveness of our approach.

The terminal Mean Absolute Error (MAE) loss achieved is depicted in Fig. \ref{MAE_Loss_Comparison}. Notably, the proposed approach has precipitated a significant diminution in the loss term. This shows that we have been able to diminish lots of the added noise to the frames due to the drop in the bitrate of the video.

Regarding the Peak Signal-to-Noise Ratio (PSNR) and Structural Similarity Index (SSIM), our video was reconstructed using the revitalized frames, followed by the computation of both evaluation metrics. Fig.\ref{PSNR_SSIM} demonstrates our achievement in meeting both SSIM and PSNR criteria by elevating the results from an unacceptable quality measure to an acceptable threshold. Specifically, acceptable PSNR values must exceed 30, a benchmark successfully met by all three proposed models. Furthermore, we have elevated SSIM results from below the acceptable threshold of 0.9 to above it, thereby aligning them within the appropriate quality range for SSIM.

\begin{figure*}[hbt!]
\centering
\includegraphics[width=6.3in]{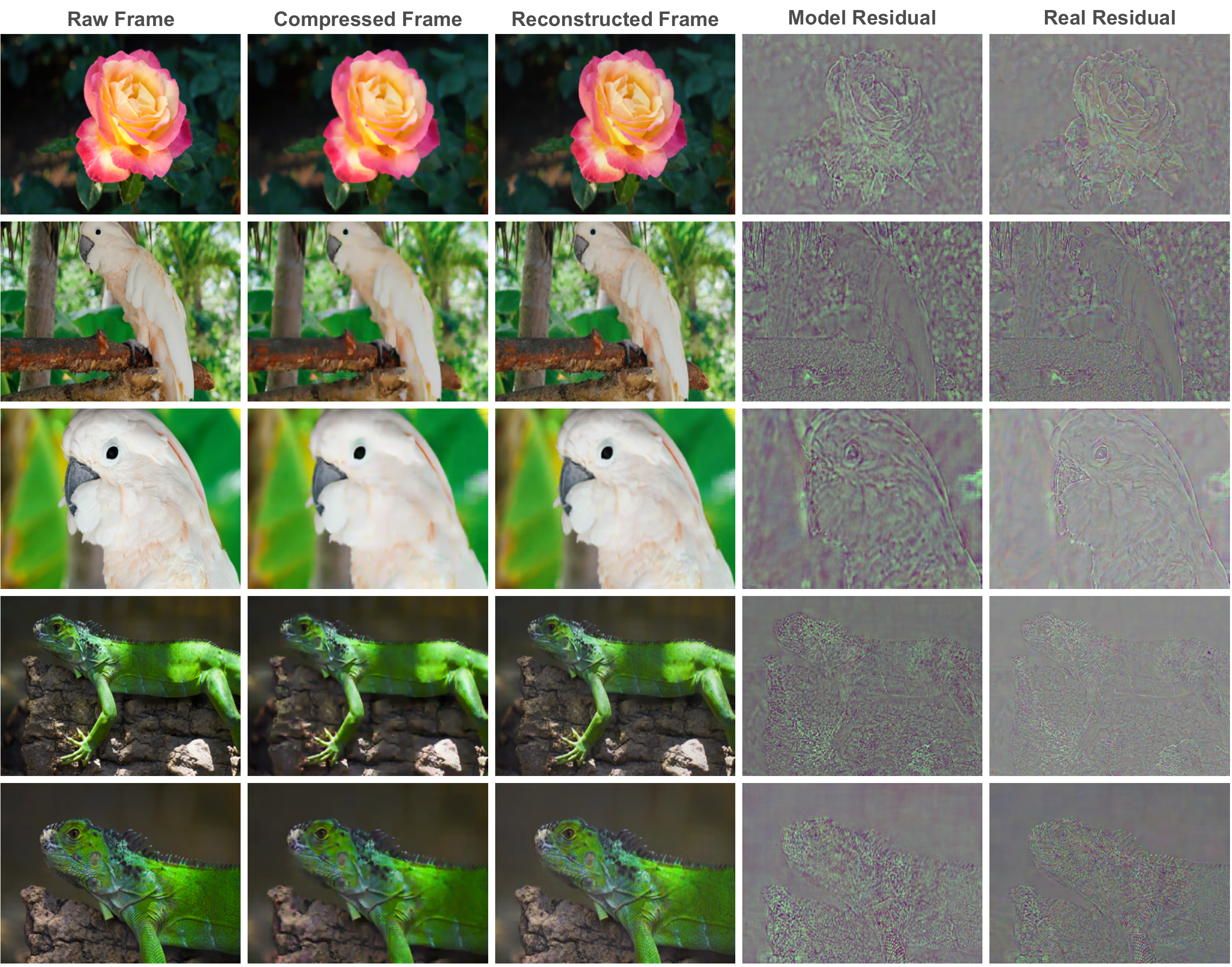}
\caption{Visualization of Frames Reconstructed by the NU-Class Net, Demonstrating Fidelity and Accuracy in Reproduction.}
\label{fig12}
\end{figure*}

Fig.\ref{fig12} exhibits a qualitative comparison across raw, compressed, and reconstructed frames, illustrated through five distinct frames within the test dataset. The figure presented here offers a distinct visualization of the effectiveness of our model in practical scenarios, particularly in enhancing the quality of video frames. It demonstrates the model's proficiency in predicting the residual of the frames, which represents the discrepancy between the original and reduced bitrate frames. This capability is shown to achieve a commendable level of accuracy in real-world applications. As depicted in Fig. \ref{fig12}, the model notably excels in restoring and augmenting the lost quality of video frames, achieving a high standard of output quality.

\textbf{Execution Time Analysis}. A pivotal consideration in our experimentation is the execution time, particularly as the Constant Rate Factor (CRF) escalates at the edge node, thereby inducing a diminished quality and bit rate, coupled with a truncated execution time, which subsequently alleviates the computational power requisitions at the edge nodes. Our assessments illuminate that amplifying the CRF from a baseline value of 18 to 40 precipitates a 63 percent diminution in the encoder's execution time. While the NU-Class Net model predominantly operates on devices possessing more lenient resource constraints compared to edge-based cameras—such as Internet of Things (IoT) server nodes, it remains imperative to finalize video frame processing within a plausible timeframe. The empirical evaluation reveals that the execution time of the NU-Class Net, with previously introduced parameters, on a V100 GPU for a singular frame, stands at 83.33ms, thereby facilitating video processing at a rate of 24 frames per second. In a bid to curtail this execution time and enable operability on more elementary CPUs and GPUs, various deep learning hardware and software acceleration methodologies can be pondered \cite{daneshtalab2020hardware}. Within the scope of this paper, parameter quantization was applied to NU-Class Net. The impact of substituting the representation of the model parameters from its innate 32-bit floating-point numbers to narrower fixed-point numbers was scrutinized. Observations underscore that attenuating the precision to 14-bit fixed-point numbers sustains the output accuracy within one percent of the original. Consequently, the model operates with 16-bit fixed-point numbers (the closest word size to 14), contracting the model size by approximately 2x and reducing the execution time to 19.84ms. It is pivotal to note that quantization is among the more rudimentary acceleration methodologies; an exploration into more advanced accelerations is deferred for future work. For example, the methodology proposed by Neda et al. \cite{NegarNeda} could be harnessed to optimize our network further.



\section{Conclusion}
\noindent In the current work, we have introduced NU-Class Net, a groundbreaking deep learning model engineered to mitigate coding noise within low-bit-rate compressed videos. Capitalizing on the robust architecture of NU-Class Net, our approach adeptly revitalizes a myriad of details that are frequently lost during video compression, without necessitating the creation of new or alterations to existing video codecs. Instead, our model strategically operates post-decoding, utilizing a conventional codec to recover the intricate video details that tend to be compromised by a low bit-rate encoder. The capabilities offered by our method are extensive, facilitating seamless integration with a wide array of existing video codecs. Evidenced by up to a 40\% improvement in perceivable video quality, as indicated by MAE Loss on our utilized benchmarks, NU-Class Net stands out as a significant contribution to the field.

The outcomes of our research serve not merely as a validation but as a tangible proof of concept, illustrating the feasibility of employing deep learning models to facilitate a high-quality video experience in IoT systems, particularly those constrained by limited computational power and bandwidth at the edge nodes. Looking forward, future research endeavors will necessitate the amplification of the neural network model's performance—potentially through the infusion of additional video-specific capabilities and refined training methodologies—and a reduction in its complexity, which may be achieved via the application of hardware acceleration. This progression will serve to further cement the applicability and efficiency of deep learning models in optimizing video quality within constrained IoT environments.

\bibliographystyle{IEEEtran}
\bibliography{Main}







\end{document}